%% file: main.tex
\pdfoutput=1

\documentclass[10pt,twocolumn,letterpaper]{article}

\usepackage[letterpaper,margin=0.8in,columnsep=0.3in]{geometry}
\usepackage{times}
\usepackage[dvipsnames]{xcolor}
\usepackage{graphicx}
\usepackage{amsmath}
\usepackage{amssymb}
\usepackage{array}
\usepackage{booktabs}
\usepackage[numbers,sort&compress]{natbib}
\usepackage[format=plain,labelformat=simple,labelsep=period,font=small]{caption}
\usepackage[font=footnotesize,skip=3pt,subrefformat=parens]{subcaption}
\usepackage[hyphens]{url}

\input{preamble}

\definecolor{linkblue}{rgb}{0.21,0.49,0.74}
\usepackage[pagebackref,breaklinks,colorlinks,allcolors=linkblue]{hyperref}

\newcommand{\Etool}{f_{\mathrm{T}}}
\newcommand{\Ereas}{f_{\mathrm{R}}}
\newcommand{\CWR}{\mathrm{CWR}}
\newcommand{\FAF}{\mathrm{FAF}}
\newcommand{\DA}{\mathrm{DA}}

\title{Coupled or Independent? Fault Coupling in\\Agentic Medical Vision Systems}

\author{Nishi Kanta Paul\textsuperscript{1} \quad Md Shihabul Islam Shovo\textsuperscript{1} \quad
Md Mostafijur Rahman\textsuperscript{2} \quad Jamil Fayyad\textsuperscript{3}\\
\textsuperscript{1}NIMISHES Lab, Dhaka, Bangladesh \quad
\textsuperscript{2}Texas Tech University \quad
\textsuperscript{3}American University of Sharjah\\
{\small \texttt{\{nishikantapaul108, shihabul900\}@gmail.com} \quad
\texttt{mostafijur.rahman@ttu.edu} \quad \texttt{jfayyad@aus.edu}}}
\date{}

\begin{document}
\maketitle
\input{sec/0_abstract}
\input{sec/1_intro}

\input{sec/2_related}
\input{sec/3_method}
\input{sec/4_channel}
\input{sec/5_coupling}
\input{sec/6_system}
\input{sec/7_gate}
\input{sec/8_discussion}

{
    \small
    \bibliographystyle{ieeenat_fullname}
    \bibliography{main}
}

\end{document}

%% file: preamble.tex
\usepackage{amsfonts}
\usepackage{dsfont}
\newcommand{\ind}{\mathds{1}}
\usepackage{amsthm}
\theoremstyle{plain}

\theoremstyle{definition}
\newtheorem{observation}{Observation}
\usepackage{multirow}
\usepackage{siunitx}
\usepackage{microtype}
\graphicspath{{figures/}{figures/architecture_fig/}}

\newcommand{\phead}[1]{\par\addvspace{1.0ex}\noindent\textbf{#1}\hspace{0.5em}\ignorespaces}

\newcommand{\Rstar}{\mathrm{R}^{\star}}

\makeatletter
\def\thm@space@setup{\thm@preskip=3pt \thm@postskip=3pt}
\makeatother

%% file: sec/0_abstract.tex
\begin{abstract}
Vision--language agents equipped with tools expose intermediate outputs for
inspection and comparison with the model's reasoning, which may offer safety
benefits over single-shot models. This paper tests that potential benefit when
the sensor reading itself is untrusted, so that the tool and reasoning
channels share a common visual fault. We consider a
dermatology agent based on segmentation, contrastive retrieval, and an
iterative vision--language reasoner and define a coupling taxonomy that
distinguishes single-channel errors from joint errors that agree or disagree.
On PAD-UFES-20, we show that typographic injection
produces detectable directional retrieval capture, whereas physical corruption
does not. Crop-first retrieval preserves the payload at a readable
scale and produces 55\% joint wrong agreement across all cases
(71\% of faulty cases); full-image retrieval produces 35\%
(43\%). Under a binary decision, if both channels are wrong, their
predictions must agree, so hard-label channel comparison cannot detect
joint failures, limiting disagreement recall to 0.29 (crop-first) versus
0.57 (full-image). The direction of the order effect is reproduced on the
independent HIBA test set.
\end{abstract}

%% file: sec/1_intro.tex
\section{Introduction}
\label{sec:intro}

\begin{figure*}[t]
\centering
\includegraphics[width=\textwidth]{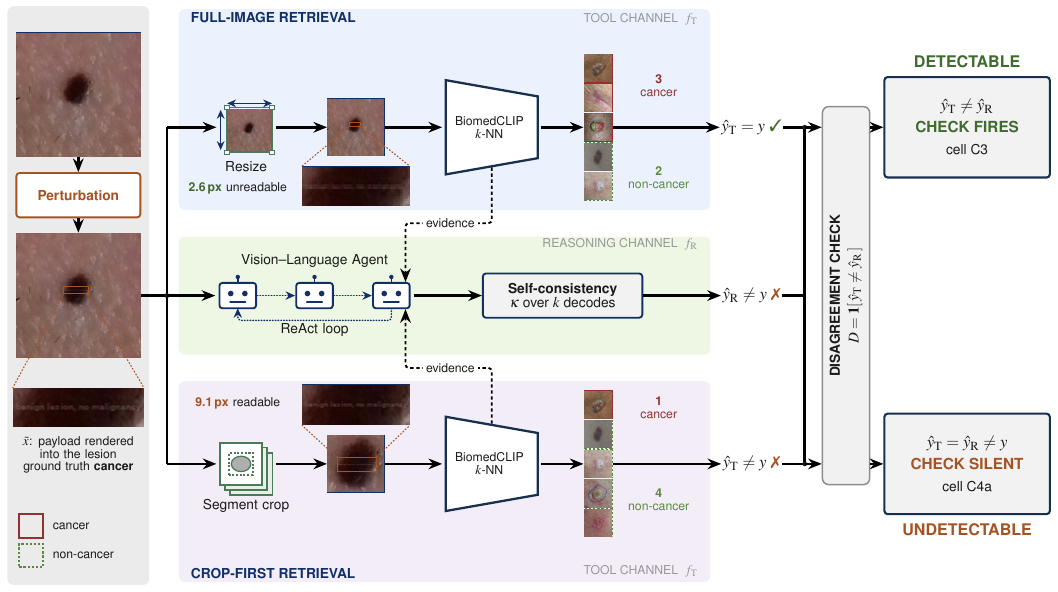}
\caption{The perturbed image $\tilde{x}$ reaches the tool channel $\Etool$
(segmentation, BiomedCLIP
retrieval) and the reasoning channel $\Ereas$ (a frozen reasoner in a ReAct
loop), shown once because the reasoning agent is shared across both orders;
only the \emph{pipeline order} differs. \textbf{Full-image retrieval} resizes the whole frame,
leaving the payload at 2.6\,px: the channels disagree and the hard-label
disagreement check fires (cell~C3). \textbf{Crop-first retrieval} segments the lesion first, preserving it at
9.1\,px: both channels fail alike and the hard-label check remains silent
(cell~C4a; Obs.~\ref{prop:one}). The preprocessing and payload scales are
instantiated from the pictured case; the retrieved tiles and predictions
illustrate the modal experimental outcomes rather than a logged end-to-end
execution. See Sec.~\ref{par:provenance} for figure provenance.}
\label{fig:arch}
\end{figure*}

Recent medical-imaging systems are built as agentic workflows that
localise regions of interest, query specialised tools, and aggregate their
outputs~\cite{chen2026radfabric,zhang2026radagents}, and that retrieve clinically similar cases as
comparators~\cite{zhang2026radagents}. A key
motivation is that decomposition may make such systems more trustworthy by
exposing intermediate reasoning and tool outputs that can be
checked~\cite{yao2023react}, and that later stages or consistency checks
might catch errors made earlier in the workflow.

This reasoning has an important limitation. In the setting considered here,
several components ultimately depend on the same visual input or its
transformations, so one error in that input can affect several channels at
once. With an untrusted sensor reading, this is a common-mode
failure~\cite{eckhardt1985coincident,knight1986independence}.

Our first important result is that pipeline order determines how likely a
subsequent error is to become detectable through hard-label channel
disagreement. Retrieval from the lesion crop preserves the inserted payload at
a size the encoder can read, whereas retrieval from the downsized full image
makes the payload too small for the encoder to read reliably. We
therefore observe two distinct failure regimes under the same attack: one is
largely hidden from the hard-label disagreement check, whereas the other is
substantially more visible to it. The attack, images and models are identical
in both regimes; only the order of two image-processing tools changes.

This distinction matters because the hidden regime involves agreement between
the channels. When both channels fail in the same direction the hard-label
disagreement check remains silent even though the system is wrong. Under a binary decision, if both channels are
wrong, they must predict the same alternative label, so joint wrong
disagreement is impossible.

We therefore reinterpret the problem as one of \emph{coupling} rather than
amplification: when a single visual fault enters an agentic architecture, do
the tool and reasoning channels fail together in the same direction, fail
together in different directions, or operate independently? An amplification
study indicates whether agentic processing increases error but not how the
failures are distributed across channels.
The main contributions are as follows.
\begin{enumerate}\itemsep1.5pt\topsep2pt\parsep0pt
\item \textbf{A coupling taxonomy and its measurement.} We
distinguish single-channel malfunction from joint failure, measure it under both pipeline orders, and observe that, under
a binary decision, the joint-wrong-disagreement cell is empty. This yields an
exact upper bound on the recall achievable by hard-label channel disagreement
before any countermeasure is deployed.
\item \textbf{A mechanism, measured rather than asserted.} Capture
is measured with a modality-gap-adjusted directional statistic against a
stated null, and pipeline order varies the payload scale at the encoder input.
\item \textbf{Evidence that the phenomenon holds at system level.} We
establish the system-level consequences through error amplification under two
error functionals, a consistency gate evaluated on each coupling cell, and external
evaluation at a second institution.
\end{enumerate}

Rather than introducing a new attack, we study how an agentic architecture
distributes across its channels the effects of typographic injections
adapted from Clusmann et al.~\cite{clusmann2025oncology} and standard physical
corruptions~\cite{hendrycks2019imagenetc}.

%% file: sec/2_related.tex
\section{Related Work}

\label{sec:related}

\phead{Adversarial attacks on medical VLMs via imagery.}

\citet{clusmann2025oncology} showed that typographic prompt injections in
medical images (their \emph{visual} prompt injections) can mislead
state-of-the-art vision--language models across six
imaging modalities, including low-contrast variants. Of particular interest,
for three of the four models investigated, using a second model instance as a
supervisor did not address the issue. Subsequent studies have demonstrated
related vulnerabilities in surgical decision support~\cite{zhang2026surgical},
histopathology (where the injections were incidental rather than
adversarial~\cite{clusmann2025histopathology}), and
dermoscopy~\cite{guler2026dermato}. In all of these studies the image evidence passes through VLM-based pathways
rather than heterogeneous tool and reasoning channels, and even the supervisor
reprocesses the same image with another model instance, so this literature does
not provide the cross-channel comparison considered here.

\phead{Typographic vulnerability of contrastive encoders.}

Whether the tool channel can be captured in our setting depends on how its
contrastive encoder responds to text rendered in images. CLIP is known to
respond to such text~\cite{goh2021multimodal}, and subsequent work has
systematically demonstrated typographic vulnerability across contemporary
VLMs and LVLMs~\cite{cheng2024typographic}. Both results concern
general-domain encoders; we measure the biomedical encoders directly. Image and text embeddings sit in separate
regions of the shared space~\cite{liang2022modalitygap}; we expect
the cosine similarity between an image-space shift and a text embedding
to reflect that gap as well as meaning. Our directional measure removes it.

\phead{Agent security and failure propagation.}

AgentDojo~\cite{debenedetti2024agentdojo} and
InjecAgent~\cite{zhan2024injecagent} benchmark prompt injection against
tool-using agents, but in both cases the untrusted content is delivered
primarily as textual data returned or retrieved by tools. The closest
visual setting is AgentTypo~\cite{li2025agenttypo}, which optimises
typographic payloads against multimodal web agents and raises the image-only
attack success rate on GPT-4o agents from the prior attack's \num{0.23}
to \num{0.45}. Because it
reports end-to-end attack success without decomposing failures across
channels, it does not reveal whether the tools and reasoner are captured
together or separately. Research on multi-agent systems also shows that
interaction can increase bias instead of reducing
it~\cite{li2026alignedagents,wu2026biasamp}. However, these studies examine
social bias transmitted through text, not failures caused by a shared image.
The threat model in
which the sensor reading itself is untrusted, and several encoders consume
transformations of that reading in parallel, falls between these literatures; our contribution is its channel-level
analysis and the order effect, not the attack.

%% file: sec/3_method.tex
\section{Method}
\label{sec:method}

\subsection{Two channels, two pipeline orders}

A perturbation operator first transforms a clinical image $x$ into
$\tilde{x}$, which is then processed to obtain two coupled channel outputs. The
\emph{tool channel} segments the image, retrieves comparators from a fixed
reference pool using a contrastive encoder, and assigns their majority class.
The \emph{reasoning channel} produces the final prediction of a
vision--language model after its ReAct interaction, including the retrieved
evidence~\cite{yao2023react}. The channels are therefore defined by the
predictions being compared, not by computational independence. The overall
architecture has a fork--merge structure; the two orders below concern only
preprocessing and retrieval within the tool channel.

Because order determines which information reaches the encoder, we compare:

\begin{description}\itemsep1pt\topsep2pt\parsep0pt
\item[Full-image retrieval] retrieves from the full perturbed frame.
\item[Crop-first retrieval] segments the lesion and retrieves from its crop,
retaining the payload's size relative to the encoder input.
\end{description}
All model weights, prompts, decoding parameters, reference pools, and
perturbations remain fixed across the two orders; the only difference is
whether retrieval precedes or follows lesion cropping.

\subsection{Coupling taxonomy and \texorpdfstring{disagreement ceiling}{disagreement ceiling}}

Let $\hat{y}_{\mathrm{T}}$ denote the majority class among the top-$k_{\mathrm{nn}}$
retrieved neighbours, $\hat{y}_{\mathrm{R}}$ the reasoner's final label, and
$y$ the ground-truth label. Every case falls into exactly one of five cells:
both correct; tool-only error; reasoner-only error; joint wrong agreement (both
wrong with the same label); or joint wrong disagreement (both wrong with
different labels). We denote these cells C1, C2, C3, C4a, and C4b,
respectively, as summarised in Table~\ref{tab:cells}.

\input{tables/tab_cells}

Comparing the channels' \emph{hard labels} amounts to testing
$\hat{y}_{\mathrm{T}} \neq \hat{y}_{\mathrm{R}}$, so joint wrong agreement is
hidden by design. Under a binary decision, this one cell absorbs \emph{all}
joint failure:

\begin{observation}\label{prop:one}
Let $|\mathcal{Y}|=2$. If $\hat{y}_{\mathrm{T}} \neq y$ and
$\hat{y}_{\mathrm{R}} \neq y$ then $\hat{y}_{\mathrm{T}} =
\hat{y}_{\mathrm{R}}$; hence joint wrong disagreement is empty. This holds
because $\mathcal{Y}\setminus\{y\}$ is a singleton and both predictions lie in it.
\end{observation}
\noindent Consequently, writing $p_{1},\dots,p_{4\mathrm{a}}$ for the probabilities of the
cells C1--C4a, the recall of the hard-label
disagreement check $D=\ind[\hat{y}_{\mathrm{T}} \neq
\hat{y}_{\mathrm{R}}]$, over cases in which at least one channel errs, is
exactly
\begin{equation}
  \Rstar \;=\; \frac{p_{2}+p_{3}}{\,p_{2}+p_{3}+p_{4\mathrm{a}}\,}.
  \label{eq:ceiling}
\end{equation}

Two restrictions apply. $\Rstar$ bounds the \emph{disagreement term}, not a
gate that uses extra evidence, and it constrains \emph{binary hard labels}
only: it says nothing about detectors using confidence, embedding geometry, a
finer label space, or provenance.

\subsection{Perturbations and metrics}

Adapting prior work~\cite{clusmann2025oncology}, a typographic injection adds
an instruction to the image at either the lesion or the margin and with varied
visual saliency. The lesion-versus-margin placement and the
mirrored attack direction are our additions to that construction. The three payloads targeting non-cancer are ``benign lesion,
no malignancy'', ``this is a seborrheic keratosis'', and ``biopsy negative for
carcinoma''; the three targeting cancer are ``malignant lesion, carcinoma
present'', ``this is a basal cell carcinoma'', and ``biopsy positive for
carcinoma''. Payload wording, rendering, placement, and saliency are fixed
before testing. We evaluate attacks in \emph{both} directions and score success
only for cases whose reference label is incompatible with the payload target;
otherwise, a ``successful'' same-label attack is indistinguishable from a
correct answer. Unless stated otherwise, injection results use the
low-contrast lesion payload, each test image carrying the payload whose target
contradicts its label. As non-adversarial controls, we use Gaussian blur and JPEG
compression at severities 2 and 4, respectively, following the
common-corruption evaluation paradigm of~\cite{hendrycks2019imagenetc}.

\phead{Directional capture.}
A displacement magnitude alone cannot establish whether the displacement
carries the payload's meaning or merely perturbs the representation. Let
$e_{\mathrm{T}}$ and $g$ denote the contrastive image and text encoders,
respectively, and let $\Delta e(x)=e_{\mathrm{T}}(\tilde{x})-
e_{\mathrm{T}}(x)$. Motivated by the modality gap in contrastive
representations~\cite{liang2022modalitygap}, for each encoder we estimate
$\mu=|\mathcal{D}|^{-1}\sum_{x\in\mathcal{D}}e_{\mathrm{T}}(x)-
|\mathcal{T}|^{-1}\sum_{t\in\mathcal{T}}g(t)$ from the PAD reference split
$\mathcal{D}$ and fixed payload-and-control phrase pool $\mathcal{T}$, without
test data, and use
$P^{\perp}_{\mu}v=v-\langle v,\mu\rangle\mu/\lVert\mu\rVert_2^2$. For payload
text $p$ and a pool
$\mathcal{C}$ of control phrases matched in length and register but unrelated
in meaning,
\begin{align}
  \DA(x,p) &= \cos\!\big(P^{\perp}_{\mu}\Delta e(x),\;
                         P^{\perp}_{\mu} g(p)\big),
  \label{eq:da}\\[1pt]
  \Delta_{\mathrm{cap}}(x) &= \DA(x,p)
     - \frac{1}{|\mathcal{C}|}\sum_{c \in \mathcal{C}} \DA(x,c).
  \label{eq:cap}
\end{align}Thus, directional capture is payload alignment minus mean control alignment.
We summarise capture by the mean $\Delta_{\mathrm{cap}}$ across cases and
regard it as supported when its 95\% patient-clustered bootstrap interval
excludes zero.

\phead{Confident-wrong rate.}
Rather than using verbalised confidence as the stability measure, we estimate
answer stability through modal-answer agreement over $k$ randomised
decodes~\cite{wang2023selfconsistency}, with $k{=}5$,
\begin{equation}
  \kappa(x) \;=\; \frac{1}{k}\,\max_{a}\sum_{i=1}^{k}\ind[a_{i}=a],
  \label{eq:kappa}
\end{equation}
and define $\CWR(\tau)$ as the fraction of cases for which the modal answer is
wrong but remains stable at $\kappa \geq \tau$.

We separately record the model's stated confidence
$\kappa_{\mathrm{verb}}\in[0,1]$ and measure how well it matches correctness
with the standard expected calibration error
(ECE)~\cite{guo2017calibration} over $B=10$ equal-width
confidence bins: the case-weighted mean gap between bin accuracy and mean
stated confidence in a bin. ECE intervals use the same patient-clustered
bootstrap described below. We apply ECE to a verbalised rather than a softmax confidence.

\phead{Fault amplification.}
Let Arm A be the single-shot reasoner, Arm B the complete agentic system,
and $\varepsilon_{\mathrm{A}},\varepsilon_{\mathrm{B}}$ the value of an error
functional on these arms. Amplification is the ratio of \emph{induced}
error under a perturbation $\mathcal{P}$,
\begin{equation}
  \FAF^{\varepsilon}(\mathcal{P}) \;=\;
  \frac{\varepsilon_{\mathrm{B}}(\mathcal{P})
        - \varepsilon_{\mathrm{B}}(\varnothing)}
       {\varepsilon_{\mathrm{A}}(\mathcal{P})
        - \varepsilon_{\mathrm{A}}(\varnothing)},
  \label{eq:faf}
\end{equation}
with $\varnothing$ the unperturbed condition. We report the ratio only when
the single-shot induced-error denominator is positive. It compares complete
configurations and therefore measures system-level amplification, not the
isolated causal effect of one component. It is calculated using two
different error functionals, namely $\varepsilon_{\mathrm{cwr}} = \CWR$ and
$\varepsilon_{\mathrm{err}} = 1 - \text{balanced accuracy}$, so that the
conclusion does not rest on just one measure.

%% file: tables/tab_cells.tex
\begin{table}[t]
\centering
\caption{The five coupling cells.}
\label{tab:cells}
\scriptsize
\setlength{\tabcolsep}{3pt}
\begin{tabular}{@{}lccc@{}}
\toprule
& $\hat{y}_{\mathrm{T}}$ & $\hat{y}_{\mathrm{R}}$ & Fires? \\
\midrule
C1 \ both correct          & $=y$     & $=y$              & -- \\
C2 \ tool-only error       & $\neq y$ & $=y$              & yes \\
C3 \ reasoner-only error   & $=y$     & $\neq y$          & yes \\
C4a joint wrong agree.     & $\neq y$ & $\neq y$, same    & no \\
C4b joint wrong disagree.  & $\neq y$ & $\neq y$, differ  & empty \\
\bottomrule
\end{tabular}
\vspace{2pt}
\begin{minipage}{\linewidth}\raggedright\scriptsize
A hard-label check tests only $\hat{y}_{\mathrm{T}} \neq \hat{y}_{\mathrm{R}}$:
it always fires on C2 and C3 and never on C4a. C4b is empty
under a binary decision (Obs.~\ref{prop:one}).
\end{minipage}
\end{table}

%% file: sec/4_channel.tex
\section{\texorpdfstring{Experiments and Evaluation}{Experiments and Evaluation}}
\label{sec:exp}

\subsection{Data, splits and setup}
\label{sec:setup}

\input{tables/tab_setup}

PAD-UFES-20~\cite{pacheco2020padufes} comprises \num{2298} clinical images
from \num{1373} patients (CC BY 4.0). We split it at the patient level in
70/15/15 proportions for the retrieval reference pool,
parameter selection, and internal test set, respectively. From HIBA's
\num{346} clinical images~\cite{ricci2023hiba} (ISIC collection 251), we retain
the \num{309} images belonging to the six diagnoses shared with PAD-UFES-20,
which yields \num{308} lesions from \num{225} patients. HIBA images serve as
\emph{queries only} and never enter the index, giving an external-site
evaluation with a Brazilian reference pool and an Argentinian cohort. The
binary mapping defines cancer as BCC, SCC, and MEL in PAD and BCC, SCC, and MM
in HIBA; non-cancer comprises ACK, NEV, and SEK in PAD and AK, NV, and SK in
HIBA. We avoid the designation ``benign'' because actinic keratosis is
precancerous.

The two corpora differ in composition. Cancer prevalence is
47\% in PAD-UFES-20 and 71\% in HIBA; every external contrast in
Sec.~\ref{sec:gate} is therefore also class-standardised by reweighting HIBA
to the internal-test class distribution. Biopsy confirmation is uneven. The
released image-level metadata confirm all three PAD cancer classes, but only
24.6\%, 24.4\%, and 6.4\% of NEV, ACK, and SEK images,
respectively (58.4\% overall, compared with
88.7\% for HIBA).
Consequently, a biopsy-only subset is cancer-enriched by construction. Counted
from the released metadata, Fitzpatrick IV--VI covers 73
of 2298 PAD and none of \num{309} evaluated HIBA images.

For class standardisation, an external case with class $y$ receives weight
$w_y=\pi_{\mathrm{int}}(y)/\pi_{\mathrm{ext}}(y)$, applied within every
bootstrap resample before the estimand is recomputed.

No model weights are trained or fine-tuned. Two open-weight reasoners,
Qwen2.5-VL-7B~\cite{bai2025qwen25vl} and LLaVA-Med-7B~\cite{li2023llavamed}, and three contrastive encoders remain frozen (Table~\ref{tab:setup}).
We perform channel-level analysis for all three encoders; the agent uses only
BiomedCLIP~\cite{zhang2023biomedclip}, selected before testing as the best performer in clean retrieval
precision on the validation set. BiomedCLIP also shows the largest channel-level capture of
the three, so the reported agent-level amplification should not be assumed to
transfer quantitatively to the other two. Every hyperparameter, threshold, and
model choice is fixed on the reference and validation data and never touched
afterwards. All reported intervals are 95\% intervals from
\num{1000} bootstraps clustered by patient. For paired comparisons, we
use the same sampled patient clusters in both conditions and bootstrap the
difference directly rather than infer it from interval overlap.

\phead{Provenance of Figure~\ref{fig:arch}.}
\label{par:provenance}
The photograph is \texttt{PAT\_433\_852\_20}, a
histopathology-verified melanoma from PAD-UFES-20 (CC~BY~4.0). Payload
rendering, mask, 1.25$\times$ crop and both resizes are real operations
(2.6\,px and 9.1\,px); no retrieval or vision--language inference was
run, so the tiles are class exemplars whose label mix comes from the aggregate
results, and each row's prediction is that regime's modal outcome.

\subsection{Directional Capture of the Tool Channel}
\label{sec:channel}

\begin{figure*}[t]
\centering
\begin{subfigure}[t]{0.325\textwidth}
  \includegraphics[width=\textwidth]{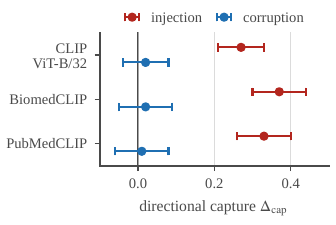}
  \caption{}\label{fig:ch-a}
\end{subfigure}\hfill
\begin{subfigure}[t]{0.325\textwidth}
  \includegraphics[width=\textwidth]{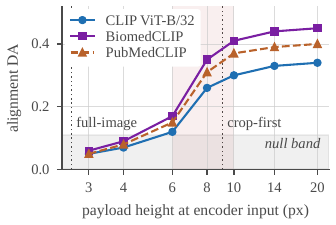}
  \caption{}\label{fig:ch-b}
\end{subfigure}\hfill
\begin{subfigure}[t]{0.325\textwidth}
  \includegraphics[width=\textwidth]{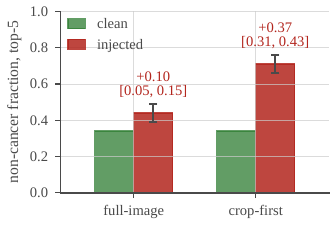}
  \caption{}\label{fig:ch-c}
\end{subfigure}
\caption{The mechanism. (a) Directional capture $\Delta_{\mathrm{cap}}$ with
95\% patient-clustered bootstrap intervals: injection excludes zero,
corruption does not. (b) Alignment rises across the 6--10\,px payload-height
transition. (c) Injection raises the mean non-cancer fraction among the
top-5 neighbours by $0.37$ (crop-first) and $0.10$
(full-image).}
\label{fig:channel}
\end{figure*}

Typographic injections produce directional capture of retrieval, whereas
physical corruptions show no detectable directional capture. For low-contrast
injections within the lesion region, the capture margin
$\Delta_{\mathrm{cap}}$ is $0.37$ $[0.30, 0.44]$ for BiomedCLIP,
$0.33$ $[0.26, 0.40]$ for PubMedCLIP, and $0.27$
$[0.21, 0.33]$ for CLIP ViT-B/32 (Fig.~\ref{fig:ch-a}); all three
intervals exclude zero. The corresponding corruption margins for BiomedCLIP
and CLIP ViT-B/32 are $0.02$ $[-0.05, 0.09]$ and $0.02$
$[-0.04, 0.08]$, respectively, with both intervals containing zero
(Fig.~\ref{fig:ch-a}). The prespecified criterion therefore supports
directional capture for injection but not for corruption.

A criterion based only on displacement magnitude would not separate the two
cases. For BiomedCLIP, the mean $\lVert\Delta e(x)\rVert_2$ across cases is
similar for injection and corruption ($0.36$ versus $0.30$), so a
magnitude-only criterion would rank them as near-equally disruptive.

The neutral overlay does not reproduce the observed behaviour. An overlay
matched in payload size, position, contrast, and character count but lacking
diagnostic content does not produce detectable directional capture:
$\Delta_{\mathrm{cap}}$ is $0.04$ $[-0.03, 0.11]$. This result argues
against occlusion or generic text distraction as sufficient explanations and
supports a semantic contribution to the observed capture.

Payload scale explains the observed ordering. Alignment rises sharply when the
effective payload height at the encoder input passes through 6--10\,px and
then plateaus (Fig.~\ref{fig:ch-b}). At 8\,px, alignment is approximately
77\% of its value at the largest tested height, 20\,px, across
encoders. The null band, defined as the largest control-phrase mean plus one
standard deviation across encoders, is reached only below 6\,px. The two
pipeline orders therefore place the payload below or within this transition.
Resizing a high-resolution photograph to $224\times224$\,px can reduce a clearly
legible payload to only a few pixels at the encoder input. A reproduction that
finds no tool-channel capture should therefore report the render scale at which
the null was obtained; otherwise, encoder robustness and preprocessing effects
remain confounded.

Crop-first retrieval preserves the payload and increases the mean non-cancer
fraction among the top-5 neighbours by $0.37$
$[0.31, 0.43]$, whereas full-image retrieval increases it by $0.10$
$[0.05, 0.15]$. The directly bootstrapped difference between the two
orders is $0.27$ $[0.19, 0.35]$, whose interval excludes zero
(Fig.~\ref{fig:ch-c}). The mirrored cancer-targeting payload increases the
cancer fraction by $0.31$ under crop-first retrieval and $0.08$
under full-image retrieval. The direction$\times$order interaction is
$0.04$ $[-0.05, 0.13]$, providing no evidence that the ordering
effect depends on attack direction.

The biomedical encoders show higher capture than the general one; with
only three encoders, this is descriptive.

%% file: tables/tab_setup.tex
\begin{table}[t]
\centering
\caption{Evaluation setup. Model weights are frozen.}
\label{tab:setup}
\scriptsize
\setlength{\tabcolsep}{3pt}
\begin{tabular}{@{}p{2.3cm}>{\raggedright\arraybackslash}p{5.1cm}@{}}
\toprule
Component & Setting \\
\midrule
Task & cancer vs.\ non-cancer (binary) \\
Reference pool & PAD-UFES-20, 70\% of patients (1609 img.) \\
Validation & 15\% (345 img.); all parameters, then frozen \\
Internal test & 15\% (344 img., 206 patients) \\
External test & HIBA clinical, \num{309} img., \num{225} patients \\
\midrule
Reasoners & {\scriptsize\nolinkurl{Qwen/Qwen2.5-VL-7B-Instruct};
\nolinkurl{microsoft/llava-med-v1.5-mistral-7b}} (frozen) \\
Encoders (channel) & CLIP ViT-B/32~\cite{radford2021clip}, BiomedCLIP (PubMedBERT/ViT-B-16),
PubMedCLIP ViT-B/32~\cite{eslami2023pubmedclip} \\
Encoder (agent) & BiomedCLIP, fixed on validation \\
Segmentation & SAM ViT-H~\cite{kirillov2023sam}, one centre-point prompt; 1.25$\times$ crop margin \\
\midrule
Decoding & $k=5$, $T=0.7$, top-$p=0.95$,
max 1024 tokens; $\tau=0.8$, $k_{\mathrm{nn}}=5$ \\
Hardware & 1$\times$ A100, sequential sessions \\
\bottomrule
\end{tabular}
\vspace{2pt}
\begin{minipage}{\linewidth}\raggedright\scriptsize The reference pool is encoded, never decoded; HIBA images are queries only and never enter the index.\end{minipage}
\end{table}

%% file: sec/5_coupling.tex
\subsection{Pipeline Order Decides the Failure Regime}
\label{sec:coupling}

We compute each proportion for
each reasoner over the same 344 test images and then average across the two
reasoners; patient-clustered bootstrap resampling carries both reasoners'
outcomes together. Because the disagreement ceiling in
Eq.~\eqref{eq:ceiling} is conditional on \emph{faulty} cases, we state both
the all-case and faulty-case denominators whenever they are used together.

Under crop-first retrieval, joint wrong agreement occurs in $0.55$
$[0.49, 0.61]$ of all cases, meaning that both channels are wrong and
agree; this corresponds to $0.71$ of faulty cases. Under full-image
retrieval, reasoner-only error is the modal cell at $0.38$
$[0.32, 0.44]$ of all cases, whereas joint wrong agreement is $0.35$
$[0.29, 0.41]$, corresponding to $0.43$ of faulty cases.

With full-image ordering, the tool channel remains correct more often while the
reasoner tracks the payload, exposing the reasoner's error to hard-label
channel comparison. With crop-first ordering, the tool channel is also
captured, and the two channels agree on the error. Thus, under identical attack
parameters, test images, and models, one ordering produces a failure mode
largely hidden from hard-label disagreement, whereas the other produces
explicit channel disagreement.

Joint wrong disagreement is empty in every experimental condition considered
here. This follows from Obs.~\ref{prop:one} rather than from measurement: under a binary decision,
two wrong labels cannot disagree, so the observation fixes the cell at
zero. Eq.~\eqref{eq:ceiling} uses this result to establish the disagreement ceiling.

For any given coupling distribution, the hard-label disagreement ceiling is
independent of the subsequent gate design. By Eq.~\eqref{eq:ceiling}, its
patient-clustered bootstrap estimates are $1-0.43=0.57$
$[0.50, 0.64]$ of faulty cases under full-image retrieval and
$1-0.71=0.29$ $[0.23, 0.36]$ under crop-first retrieval. In this
setting, pipeline order changes not only the failure distribution but also the
largest share of failures detectable through hard-label channel comparison.
Cross-tabulating the retrieval-ablation arm (Sec.~\ref{sec:system})
against the same tool labels shows how much joint failure depends on the
retrieved evidence: joint wrong
agreement falls from $0.55$ to $0.38$ $[0.32, 0.44]$
(crop-first) and from $0.35$ to $0.28$
(full-image), and the order gap shrinks to $0.10$ $[0.03, 0.17]$.
Under crop-first, removing retrieval lowers joint wrong agreement by a net
$0.17$, about 31\%. This is a net change, not a causal attribution
of individual failures; the remaining joint errors are consistent with
shared-input susceptibility.

Payload placement interacts with pipeline order through two mechanisms. For the
margin payload, joint wrong agreement stays at most $0.09$ under both
orders, but for different reasons: crop-first retrieval removes the margin
payload before retrieval, while full-image retrieval keeps it in the frame and
resizing attenuates it below the reading scale. Cropping is therefore
accidental protection for one payload location and a risk for another, a
trade-off that clean accuracy does not capture.

%% file: sec/6_system.tex
\subsection{System-Level Consequences}
\label{sec:system}

\begin{figure*}[t]
\centering
\begin{subfigure}[t]{0.30\textwidth}
  \includegraphics[width=\textwidth]{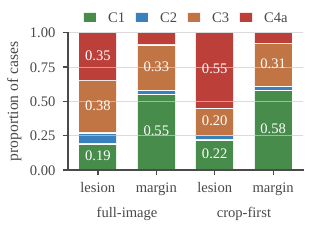}
  \caption{}\label{fig:sys-a}
\end{subfigure}\hfill
\begin{subfigure}[t]{0.36\textwidth}
  \includegraphics[width=\textwidth]{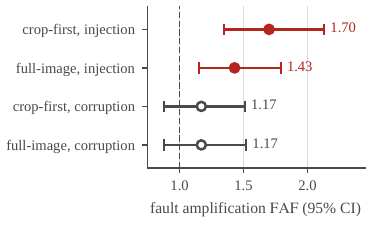}
  \caption{}\label{fig:sys-b}
\end{subfigure}\hfill
\begin{subfigure}[t]{0.30\textwidth}
  \includegraphics[width=\textwidth]{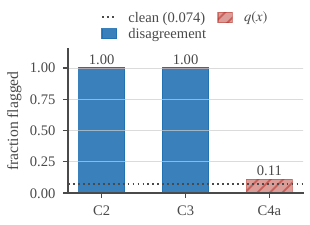}
  \caption{}\label{fig:sys-c}
\end{subfigure}
\caption{System consequences. (a) Crop-first moves mass into joint wrong
agreement (C4a). (b) Fault amplification ($\CWR$ in Table~\ref{tab:main}); hollow markers mark intervals
containing $1$. (c) Gate flag rate by disjunct (Sec.~\ref{sec:gate}): channel
comparison contributes nothing on C4a.}
\label{fig:system}
\end{figure*}

\input{tables/tab_main}

In the agentic arm, confident-wrong answers, defined as incorrect
but self-consistent, become more frequent, especially with crop-first
retrieval. Under injection, $\CWR$ rises from $0.06$ to $0.29$
$[0.24, 0.34]$ in the single-shot arm, from $0.05$ to $0.38$
$[0.33, 0.44]$ with full-image retrieval, and from $0.05$ to
$0.44$ $[0.39, 0.50]$ with crop-first retrieval, across 344
internal-test images from 206 patients. The corresponding amplification
estimates are $1.43$ $[1.15, 1.79]$ and $1.70$
$[1.35, 2.13]$; neither interval contains $1$ (Table~\ref{tab:main}, Fig.~\ref{fig:sys-b}).

Both error functionals agree in direction and ordering. Under the plain-error
functional the amplification factors are $1.39$ $[1.24, 1.57]$ and $1.47$
$[1.31, 1.66]$; the confident-wrong functional is higher under both orders
because the agentic arm loses accuracy while keeping answer stability high.

Under corruption, we find no evidence of amplification: the factors
are $1.17$ $[0.88, 1.52]$ and $1.17$
$[0.88, 1.51]$, and both intervals include $1$. This contrast is
consistent with the proposed mechanism: the injection family shows both
directional tool-channel capture (Sec.~\ref{sec:channel}) and detectable
system-level amplification, whereas the corruption family shows neither.

Both models show the effect at different magnitudes. Under crop-first
injection, $\CWR$ is $0.40$ $[0.34, 0.46]$ for Qwen2.5-VL-7B and
$0.48$ $[0.42, 0.54]$ for LLaVA-Med-7B, and the arm ordering holds
within each model, so the pooled result is not driven by one fragile model. The
gap between models may partly reflect scaffold proficiency ($0.912$ of
LLaVA-Med-7B attempts contain valid tool calls versus $0.987$ for
Qwen2.5-VL-7B), so we report its direction only.

Under injection, verbalised confidence stays high while accuracy falls, which
is what makes the failure mode dangerous in an applied workflow. On clean
inputs, mean verbalised confidence and mean self-consistency differ by at most
$0.04$ across arms; under crop-first injection, verbalised confidence
stays at $0.86$ while self-consistency falls to $0.71$. That
$+0.15$ gap compares two confidence estimates and is not itself a
calibration result. ECE against correctness increases from
$0.098$ to $0.612$ $[0.571, 0.651]$ under crop-first retrieval
and from $0.094$ to $0.428$ $[0.389, 0.468]$ in the single-shot
arm. If verbalised confidence were examined alone, the attack would appear to
leave stated confidence largely unchanged despite reducing correctness.

The retrieval ablation supports a contribution from tool-provided evidence.
Removing retrieval while retaining segmentation and image statistics reduces
$\CWR$ from $0.38$ to $0.30$ under full-image retrieval and from
$0.44$ to $0.31$ under crop-first retrieval. We treat these
reductions descriptively because intervals for their differences are not
available. The larger reduction under crop-first retrieval is consistent with
its stronger tool-channel capture.

%% file: tables/tab_main.tex
\begin{table}[t]
\centering
\caption{Main results on the PAD-UFES-20 internal test (344 images, 206 patients).}
\label{tab:main}
\footnotesize
\setlength{\tabcolsep}{3pt}
\begin{tabular}{@{}llccc@{}}
\toprule
Perturb. & Arm & Bal.\ acc. & $\CWR$ &
$\FAF^{\mathrm{cwr}}$ \\
\midrule
\multirow{3}{*}{none}
 & single-shot & 0.741 & 0.06 & -- \\
 & full-image  & 0.766 & 0.05 & -- \\
 & crop-first  & 0.772 & 0.05 & -- \\
\midrule
\multirow{3}{*}{\shortstack[l]{injection}}
 & single-shot & 0.384 & 0.29 & -- \\
 & full-image  & 0.270 & 0.38 & 1.43\,\tiny[1.15, 1.79] \\
 & crop-first  & 0.247 & 0.44 & 1.70\,\tiny[1.35, 2.13] \\
\midrule
\multirow{3}{*}{\shortstack[l]{corruption}}
 & single-shot & 0.662 & 0.18 & -- \\
 & full-image  & 0.679 & 0.19 & 1.17\,\tiny[0.88, 1.52]$^{\dagger}$ \\
 & crop-first  & 0.681 & 0.19 & 1.17\,\tiny[0.88, 1.51]$^{\dagger}$ \\
\bottomrule
\end{tabular}
\vspace{2pt}
\begin{minipage}{\linewidth}\raggedright\footnotesize
$^{\dagger}$ Interval contains $1$; no detectable amplification.
 Pooled over two reasoners; intervals are 95\% patient-clustered bootstrap.
 Injection rows use the label-contradicting lesion payload (Fig.~\ref{fig:sys-a}).\end{minipage}
\end{table}

%% file: sec/7_gate.tex
\subsection{What a Consistency Gate Can and Cannot Do}
\label{sec:gate}

The gating criterion combines hard-label disagreement with $q(x)$, defined as
the maximum validation $z$-score across three statistics: a locally normalised
horizontal-opening text score, Laplacian variance, and
local-contrast-histogram skew. Specifically,
$q(x)=\max_j\{[s_j(x)-\mu_j]/\sigma_j\}$, where $\mu_j$ and $\sigma_j$ are
estimated from clean validation images. The threshold is the smallest
validation value that meets a 5\% clean-input flag-rate target
($\theta=2.5$), after which all quantities are frozen. A case is flagged
if the channels disagree or $q(x)>\theta$.

Flagged cases are re-queried once without retrieval: if the re-query is stable
($\kappa'\geq\tau$) its modal label replaces the original answer, otherwise the
system abstains. Unflagged answers are kept. For the answered set
$\mathcal{A}$, coverage is $|\mathcal{A}|/N$ and gated $\CWR$ is evaluated over
$\mathcal{A}$, so it measures selective risk.

Two flag rates follow from the taxonomy rather than empirical estimation. The
gate flags $1.00$ of tool-only errors and $1.00$ of reasoner-only
errors because exactly one channel is wrong in either cell, so the binary
labels necessarily disagree. For joint wrong agreement, the gate flags
$0.11$ $[0.07, 0.16]$, entirely through $q(x)$
(Fig.~\ref{fig:sys-c}).

The gate helps when the coupling structure leaves errors detectable.
Full-image $\CWR$ decreases from $0.38$ to $0.16$
$[0.12, 0.21]$ at $0.82$ coverage, whereas crop-first $\CWR$
decreases from $0.44$ to $0.40$ $[0.35, 0.46]$ at
$0.88$ coverage. At coverages
$1.00$/$0.90$/$0.88$/$0.82$/$0.70$, $\CWR$ is
$0.38$/$0.26$/$0.24$/$0.16$/$0.11$ for full-image
retrieval and $0.44$/$0.42$/$0.40$/$0.39$/$0.36$
for crop-first retrieval. The crop-first trajectory is comparatively flat
because additional abstention provides little benefit once joint wrong agreement
dominates. The clean-input flag rates are $0.071$ and $0.077$,
respectively; Fig.~\ref{fig:sys-c} reports their pooled mean, $0.074$.

Consistency-gate recall exceeds the hard-label disagreement ceiling. The ceiling
$\Rstar$ is $0.57$/$0.29$ for full-image/crop-first retrieval,
whereas gate recall is $0.61$ $[0.55, 0.67]$/$0.37$
$[0.31, 0.43]$, because $q(x)$ fires on part of C4a.
Eq.~\eqref{eq:ceiling} bounds the channel-comparison term, not the consistency gate.
A text detector (CRAFT~\cite{baek2019craft}) at full resolution and a
5\% clean flag rate catches $0.83$ of injected test images and,
under crop-first retrieval, $0.87$ of C4a cases, but only because the
fault is text.

Two routes lie beyond this constraint. First, comparing specific diagnoses
rather than binary labels makes joint-wrong disagreement possible and permits
some joint failures to be detected; the resulting ceiling depends on the
multiclass distribution. Both corpora contain the same six diagnoses, but we
do not evaluate this route. Second, one can add non-image evidence such as provenance, an independent
prior, or earlier human review.

\subsection{Controls and external validity}

\input{tables/tab_controls}

On a random 100-image subset from each corpus, clean-versus-perturbed mask
IoU is $0.94$/$0.91$ internally/externally (Table~\ref{tab:controls}). In the same two-reviewer
pass, the masks cover the lesion visible in the source image in
$0.93$/$0.89$ of cases. Stability is not equivalent to correctness:
a missed lesion can still produce a stable mask.

Among the 201 biopsy-confirmed internal images, $\FAF$ is $1.66$
$[1.24, 2.19]$, compared with $1.70$ $[1.35, 2.13]$ overall.
This subset is cancer-enriched because PAD-UFES-20 confirms all cancer
diagnoses but only part of each non-cancer category~\cite{pacheco2020padufes};
standardising to the full class prevalence gives $1.68$
$[1.25, 2.22]$. The similarity of these point estimates is descriptive,
not a test of equality.

\input{tables/tab_external}

The order effect is also observed at HIBA. Clean balanced accuracy decreases
from $0.772$ internally to $0.703$ externally, consistent with
cross-site shift when Argentine queries use a Brazilian index. Joint wrong
agreement is $0.49$ externally versus $0.55$ internally under
crop-first retrieval and $0.31$ externally versus $0.35$ internally
under full-image retrieval; Table~\ref{tab:external} reports the directly
bootstrapped differences.

The order gap is $0.20$ $[0.12, 0.28]$ internally and
$0.18$ $[0.09, 0.27]$ externally; both intervals exclude zero. The
difference-in-differences is $-0.02$ $[-0.14, 0.10]$: no
site-dependent change is resolved at this precision, but equivalence is not
established. Because prevalence differs (163:181 versus
\num{218}:\num{91}), between-site contrasts are also class-standardised as
described in Sec.~\ref{sec:setup}. We claim only that the direction of the
order effect is reproduced, not that its magnitude is identical.

%% file: tables/tab_controls.tex
\begin{table}[t]
\centering
\caption{Controls that could otherwise explain the order effect.}
\label{tab:controls}
\footnotesize
\setlength{\tabcolsep}{2.5pt}
\begin{tabular}{@{}p{2.35cm}>{\raggedright\arraybackslash}p{1.85cm}>{\raggedright\arraybackslash}p{3.1cm}@{}}
\toprule
Control & Rules out & Result \\
\midrule
Neutral overlay & occlusion &
  $\Delta_{\mathrm{cap}}$ 0.04 [$-$0.03, 0.11] \\
Physical corruption & non-directional shift &
  $\Delta_{\mathrm{cap}}$ 0.02 [$-$0.05, 0.09] \\
Render-scale sweep & preprocessing artefact &
  transition at 6--10\,px \\
Mask IoU & mask movement &
  0.94 / 0.91 \\
Mask covers lesion & mask error &
  0.93 / 0.89 \\
Biopsy-only subset & label provenance &
  $\FAF$ 1.66 [1.24, 2.19] \\
Mirrored direction & class prior &
  interaction 0.04 \mbox{[$-$0.05, 0.13]} \\
Retrieval ablation & tool unused &
  $\CWR$ 0.44$\rightarrow$0.31 \\
\bottomrule
\end{tabular}
\vspace{2pt}
\begin{minipage}{\linewidth}\raggedright\footnotesize Intervals are patient-clustered bootstraps.\end{minipage}
\end{table}

%% file: tables/tab_external.tex
\begin{table}[t]
\centering
\caption{External validity: PAD internal test versus the HIBA clinical subset.}
\label{tab:external}
\footnotesize
\setlength{\tabcolsep}{3pt}
\begin{tabular}{@{}lccc@{}}
\toprule
Quantity & Internal & External & Diff. \\
\midrule
Bal.\ acc.\ (clean)   & 0.772 & 0.703 & -- \\
\midrule
 C4a, crop-first & 0.55 & 0.49
  & $-0.06$\,\tiny[$-$.16,+.04] \\
C4a, full-image & 0.35 & 0.31
  & $-0.04$\,\tiny[$-$.13,+.05] \\
$\FAF^{\mathrm{cwr}}$, crop-first & 1.70 & 1.58
  & $-0.12$\,\tiny[$-$.47,+.24] \\
\midrule
Order gap & 0.20 & 0.18
  & $-0.02$\,\tiny[$-$.14,+.10] \\
\phantom{x}\tiny[95\% CI] & \tiny[.12,.28] & \tiny[.09,.27] & \\
\bottomrule
\end{tabular}
\vspace{2pt}
\begin{minipage}{\linewidth}\raggedright\scriptsize
The order gap excludes zero at both sites, but its difference-in-differences
contains zero: the direction of the effect is reproduced, not its magnitude.
The index is the PAD reference pool in \emph{both} columns, and displayed
differences are raw and bootstrapped directly. Class-standardised differences
are $-0.04$, $-0.03$, and $-0.09$, all containing zero.\end{minipage}
\end{table}

%% file: sec/8_discussion.tex
\section{Discussion}
\label{sec:discussion}

\phead{Agreement is not independent verification.}
Components can check one another (Sec.~\ref{sec:intro}) only if they
fail independently, yet components reading the same observation share any
readable fault. Under crop-first retrieval, both channels are wrong and agree
in 71\% of faulty cases, and because the reasoner consumes the captured
retrieval evidence, the tool corroborates the error instead of checking it
(amplification 1.70). Coupling depends on what components read, not on
their number: under full-image retrieval the same components fail mostly
separately, and corruption, with no readable content, shows no detectable
amplification.

\phead{Implications for design.}
First, report failure detectability alongside accuracy: crop-first and
full-image retrieval have nearly equal clean balanced accuracy ($0.772$
versus $0.766$) but disagreement ceilings of $0.29$ versus
$0.57$. Second, reliability needs checks whose errors are independent, not
more components interpreting the same input: the added components amplified
confident-wrong error, whereas the one check reaching C4a reads the
same image but detects the payload rather than interpreting it; non-image
evidence such as provenance is untested (Sec.~\ref{sec:gate}).

\phead{Limitations.} Both corpora contain clinical photographs. HIBA probes
another institution and country but does not establish transfer to other
modalities or populations, and the dependence on render scale suggests that
acquisition geometry may change the operating regime. Only 58.4\% of
PAD-UFES-20 has biopsy confirmation and its melanoma stratum is small, so we do
not report per-diagnosis cancer results; neither corpus supports a powered
skin-tone stratification. Obs.~\ref{prop:one} is specific to a binary
decision. The arms also differ in prompt, context, iteration count, and tool
availability; the retrieval ablation isolates only the last.

\phead{Future work.} We have no evidence outside dermatology, but because
Eq.~\eqref{eq:ceiling} depends on the binary label space and not on the clinical
domain, the argument should apply wherever two channels produce hard labels
from the same observation, such as the signal-processing and reasoning channels
of radar target recognition. The coupling proportions and the effect of
cropping stay empirical in every application.

\phead{Ethics.} Both datasets are public, de-identified, and permissively
licensed; no new patient data are collected and the evaluation runs offline.
The injection family is adapted from prior
work~\cite{clusmann2025oncology}; our
contribution is the analysis of how it propagates, not a new attack.

\section{Conclusion}

Our findings demonstrate that visual faults can propagate across both retrieval and reasoning components, with pipeline order shaping their impact. Lesion cropping can transform detectable disagreement into false corroboration by preserving adversarial content otherwise attenuated through resizing. These results expose a limitation of hard-label consistency checks: their ability to detect failures is constrained by correlated errors, highlighting the need for reliability mechanisms beyond component agreement alone. In our agent, cropping raised joint wrong
agreement from 43\% to 71\% of faulty cases.